\documentclass[10pt,twocolumn,letterpaper]{article}

\usepackage{cvpr}              
\definecolor{cvprblue}{rgb}{0.21,0.49,0.74}
\usepackage[pagebackref,breaklinks,colorlinks,allcolors=cvprblue]{hyperref}

\usepackage{adjustbox}
\usepackage{multirow}
\usepackage{pgfplots} 
\usepackage{hyperref}

\def\paperID{12411} 
\def\confName{CVPR}
\def\confYear{2026}

\title{Gaussian Shannon: High-Precision Diffusion Model Watermarking  Based on Communication}

\author{
    Yi Zhang,
	Hongbo Huang,
	Liang-Jie Zhang\thanks{Corresponding author}
	\\
	Center for AI Services Computing, College of Computer Science and Software Engineering
	\\Shenzhen University
	\\
	{\tt\small rambo.ai@szu.edu.cn  \space hhwaves@163.com \space  zhanglj@ieee.org}
}

\begin{document}
\maketitle
\begin{abstract}
Diffusion models generate high-quality images but pose serious risks like copyright violation and disinformation. Watermarking is a key defense for tracing and authenticating AI-generated content. However, existing methods rely on threshold-based detection, which only supports fuzzy matching and cannot recover structured watermark data bit-exactly—making them unsuitable for offline verification or applications requiring lossless metadata (e.g., licensing instructions). To address this problem, in this paper, we propose \textbf{Gaussian Shannon}, a watermarking framework that treats the diffusion process as a noisy communication channel and enables both robust tracing and exact bit recovery. Our method embeds watermarks in the initial Gaussian noise without fine-tuning or quality loss. We identify two types of channel interference—local bit flips and global stochastic distortions—and design a cascaded defense combining error-correcting codes and majority voting. This ensures reliable end-to-end transmission of semantic payloads. Experiments across three Stable Diffusion variants and seven perturbation types show that Gaussian Shannon achieves state-of-the-art bit-level accuracy while maintaining a high true positive rate, enabling trustworthy rights attribution in real-world deployment. The source
code have been made available at: \url{https://github.com/Rambo-Yi/Gaussian-Shannon.git}

\end{abstract}    
\section{Introduction}
\label{sec:intro}

\begin{figure}[h]
	\centering
    \setlength{\abovecaptionskip}{3pt}
    \setlength{\belowcaptionskip}{-15pt} 
	\includegraphics[width=1\textwidth,
		trim=660pt 418pt 100pt 435pt,
		clip
	]{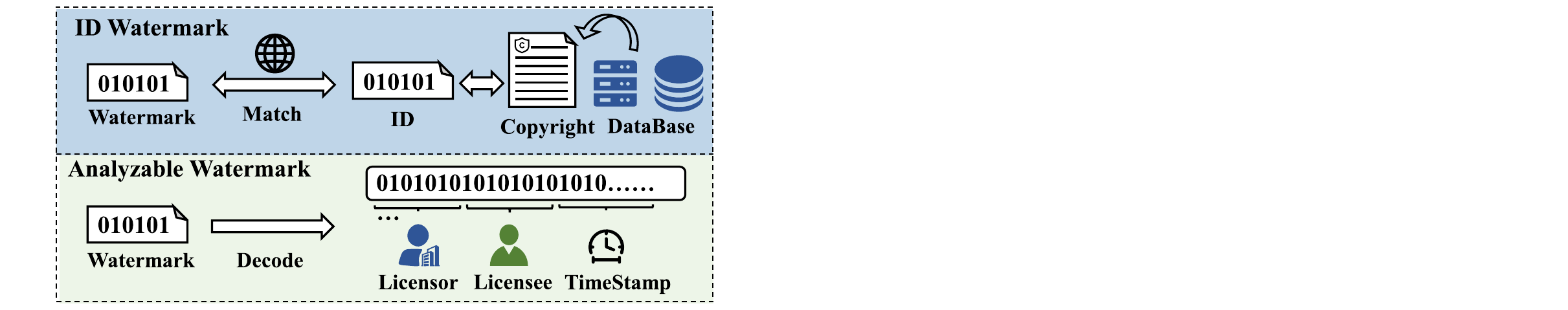}
    \caption{\textbf{Comparison of Watermark Types}. ID-based watermarks require an online connection to query a database for copyright information, whereas analytical watermarks can be directly decoded and interpreted without external resources. For example, a watermark in a digital work can contain structured data such as licensor, licensee, timestamp, and permission flags.}
	\label{fig:fig1}
\end{figure}

In recent years, diffusion models~\cite{ho2020denoising,rombach2022high,song2020denoising} have achieved remarkable success in image generation. By iteratively denoising random noise, they produce highly realistic and high-fidelity images, establishing themselves as a cornerstone of modern generative artificial intelligence. However, this powerful capability also introduces significant security risks—ranging from copyright infringement and the dissemination of disinformation to the generation of malicious content~\cite{humphreys2024ai}. For instance, adversaries have exploited diffusion models to synthesize illegal imagery for financial gain~\cite{brundage2018malicious}, undermining digital content governance and intellectual property protection.

\begin{figure*}[h]
	\centering
	\setlength{\abovecaptionskip}{1pt} 
	\setlength{\belowcaptionskip}{-15pt} 
	\includegraphics[width=0.9\textwidth,
	trim=60pt 190pt 140pt 180pt,
	clip
	]{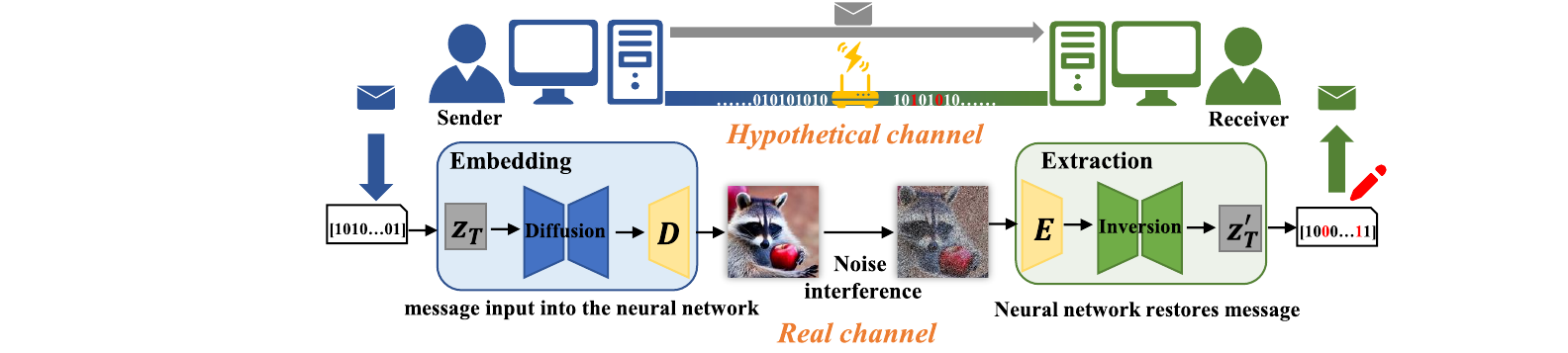}
    \caption{\textbf{Modeling Watermarking as a Communication Process.}The embedding and extraction of watermark information can be formulated as the transmission and reception of messages through a noisy channel. This perspective enables the application of established communication-theoretic techniques to enhance the reliability and fidelity of watermark recovery.}
	\label{fig:fig2}
\end{figure*}

To address these concerns, watermarking—serving as an active content protection mechanism—has been integrated into diffusion models. Recent approaches fall into two main categories. The first employs fine-tuning: one line of work trains the generative model on a watermarked dataset so that all generated images inherently carry copyright information~\cite{liu2025harnessing}; another embeds watermarks in the latent space by fine-tuning the VAE decoder to inject invisible signals during latent-to-pixel reconstruction~\cite{fernandez2023stable,kim2024wouaf}. However, both strategies incur non-negligible training or computational overhead.
To avoid fine-tuning, Tree-Ring~\cite{wen2023tree} embeds watermarks in the Fourier domain of standard Gaussian noise, though this constrains the randomness of the sampling process. GaussianShading~\cite{yang2024gaussian}, a recent fine-tuning-free approach, overcomes this limitation via watermark randomization and distribution-preserving sampling, achieving strong robustness.
More recently, PRCW~\cite{gunn2024undetectable} further improves watermark capacity and resilience through pseudo-random error-correcting codes.

Admittedly, current watermark detection mechanisms based on threshold matching have proven effective for conventional image processing and enable robust content tracing. However, such fuzzy matching is ill-suited for applications demanding bit-level fidelity of the embedded information. When batch-wise offline verification is required—or when the watermark itself encodes structured executable data (e.g., licensing instructions)—threshold-based approaches become inadequate. As illustrated in Fig.~\ref{fig:fig1}, a digital work may embed not merely a simple identifier, but a complete structured record comprising fields such as creator, timestamp, usage permissions, and cryptographic verification markers. In such cases, every bit of the watermark directly informs the final authorization decision, making lossless recovery and bitwise accuracy strict requirements. Consequently, an ideal watermarking system should operate at two levels: (1) leveraging the robustness of threshold-based matching for coarse-grained identity tracing. (2) enabling precise, full-bit extraction for reliable rights assertion in AI-generated content verification.

To achieve this objective—robust traceability and lossless information recovery—we propose \textbf{Gaussian Shannon}, a watermarking framework that formulates the embedding and extraction pipeline as a reliable communication system over a noisy channel. As illustrated in Fig.~\ref{fig:fig2}. The watermark is represented as a binary message stream transmitted through the generative process.
Since our method builds upon DDIM inversion~\cite{song2020denoising}, this “channel” corresponds to the input–output mapping of the diffusion model, where distortions from model prediction errors and adversarial image perturbations act as additive noise. To combat both localized bit flips and global stochastic disturbances caused by communication errors, we design a cascaded watermarking architecture that integrates majority voting with error-correcting codes (ECC).
Specifically, during embedding, the original bitstream undergoes ECC encoding; the resulting codeword is redundantly mapped into the latent space and further processed via pseudo-random modulation to preserve the standard Gaussian prior. At extraction, we recover the latent representation using DDIM inversion, segmentally demodulate it back to a bitstream, and apply majority voting followed by ECC decoding to reconstruct the watermark with high fidelity.

The contributions of this work are summarized as follows: 1) We identify a critical limitation in existing diffusion model watermarking—namely, the inability to ensure bitwise integrity of structured copyright metadata—and formalize the need for a dual-objective framework that simultaneously supports robust traceability and lossless information recovery. 2) We propose Gaussian Shannon, the first watermarking framework that models embedding and extraction as a reliable communication process over a noisy channel. We integrate error-correcting codes, majority voting, and Gaussian-preserving modulation to guarantee complete and accurate restoration of the semantic watermark payload while preserving generation quality. 3) Through extensive experiments across three Stable Diffusion variants and seven perturbation types, we demonstrate that Gaussian Shannon achieves state-of-the-art performance in both detection robustness and bit-level fidelity, enabling practical offline verification and trustworthy rights attribution in real-world scenarios.

\section{Related Work}
\label{sec:formatting}
In this section, we review related works on diffusion models and image watermarking for diffusion models.

\begin{figure*}[h]
	\centering
    \setlength{\abovecaptionskip}{1pt} 
	\setlength{\belowcaptionskip}{-15pt} 
	\includegraphics[width=0.8\textwidth,
	trim=100pt 130pt 130pt 120pt,
	clip
	]{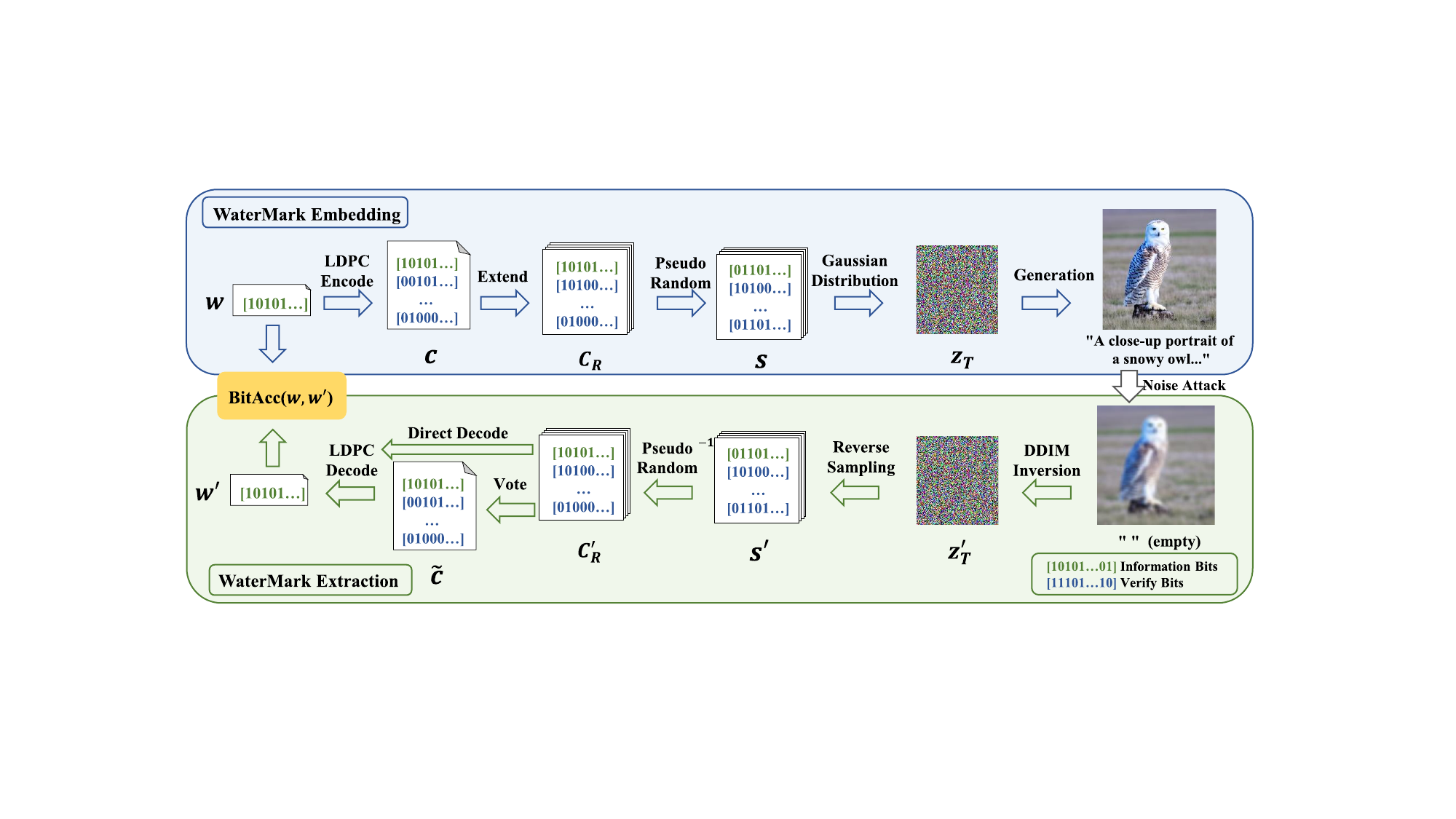}
    \caption{\textbf{Overview of the Gaussian Shannon framework. }The watermark bitstream \( \mathbf{w} \) is first encoded via LDPC into a codeword \( \mathbf{c} \), which is then expanded to match the latent space dimension to yield \( \mathbf{c}_R \). A pseudo-random modulation produces the signal \( \mathbf{s} \), which guides the sampling of the initial Gaussian noise \( \mathbf{z}_T \). The diffusion model subsequently denoises \( \mathbf{z}_T \) to generate the watermarked image. During extraction, the process is inverted to recover \( \mathbf{s}' \), followed by derandomization to obtain \( \mathbf{c}'_R \). Watermark recovery proceeds in two stages: (i) direct LDPC decoding of individual codewords in \( \mathbf{c}'_R \), or (ii) majority voting across redundant codewords to form an aggregated codeword \( \tilde{\mathbf{c}} \), which is then decoded to reconstruct \( \mathbf{w} \).
}
	\label{fig:fig3}
\end{figure*}

\subsection{Diffusion models}


Although the concept of diffusion models was originally proposed by Sohl-Dickstein et al. in 2015, the representative work by Ho et al. in 2020, Denoising Diffusion Probabilistic Models (DDPM) \cite{ho2020denoising}, achieved high-quality image generation by defining a Markov chain for forward noising and reverse denoising. Compared to other generative models like GANs\cite{goodfellow2014generative} and VAEs\cite{kingma2013auto}, diffusion models offer advantages such as better generation diversity and a lower tendency for mode collapse. However, their direct operation in pixel space is computationally intensive and memory-consuming.The emergence of Stable Diffusion \cite{rombach2022high}, which compresses the diffusion process into the latent space, has significantly reduced computational costs and promoted the widespread adoption of text-to-image applications.

However, generating a single image with DDPM often requires many hundreds of denoising steps. To accelerate this time-consuming sampling process, Song et al. proposed Denoising Diffusion Implicit Models (DDIM) \cite{song2020denoising}. By redesigning the probabilistic transition mechanism of the diffusion process, DDIM liberates the sampling trajectory from the constraint of following the Markov chain, allowing the model to synthesize high-quality images in fewer steps. In addition, DDIM has the characteristic of deterministic sampling, which can construct a reversible deterministic mapping.  This enables DDIM Inversion\cite{song2020denoising}, a process that can reverse a real image back into a noisy latent variable, thereby providing crucial support for tasks such as image editing.

\subsection{Image Watermarking for Diffusion Models}

Digital watermarking\cite{hur2024latent} is a technique that embeds information such as copyright into digital media.With the widespread application of generative models, existing methods mainly encode copyright information as bit sequences and embed them into the generation process based on the principles of steganography\cite{hur2024latent}. According to the location where the watermark is injected into the model, they can be classified into the following categories:

Kim and Min et al.\cite{meng2025latent,fernandez2023stable} directly modify the latent space information and fine - tune the decoder, respectively, so that the generated images are watermarked at the latent space stage.
The iterative nature of the reverse diffusion process in LDM\cite{rombach2022high} provides an opportunity for gradual watermark embedding. Liu and G. H. \cite{liu2023mirror}add watermarks step - by - step in each denoising step.
Liu et al.\cite{liu2025harnessing} fine - tunes the LDM model to integrate watermark features with the dataset, enabling the model itself to learn to generate images containing watermark signals. However, this approach introduces additional computational overhead and parameter modifications.
Li and Yang et al.\cite{pmlr-v267-li25ae,yang2024gaussian} choose to start from the initial noise. The former injects watermark information into both the spatial and frequency domains, while the latter probabilistically maps watermark bit information to the noise of a standard Gaussian distribution.


\section{Method}
In this section, we present our proposed method of high-precision diffusion model watermarking method
based on communication
in detail.

\subsection{Method overview}

Fig.~\ref{fig:fig3} illustrates the overall workflow of our proposed Gaussian Shannon framework.

\textbf{Embedding}. The binary watermark \( \mathbf{w} \) is first encoded via LDPC into a codeword \( \mathbf{c} \). To ensure robustness, \( \mathbf{c} \) is redundantly expanded to match the dimensionality of the diffusion latent space, yielding \( \mathbf{c}_R \). A pseudo-random modulation is then applied to produce a signal \( \mathbf{s} \) that preserves the standard Gaussian prior. Finally, the watermarked image is generated through the standard diffusion sampling process, initialized with noise shaped by \( \mathbf{s} \).

\textbf{Extraction}. Given a potentially perturbed image (e.g., after online sharing), we first apply DDIM inversion~\cite{song2020denoising} with an empty prompt to recover the initial noise estimate. This latent representation is demodulated to obtain multiple noisy copies of the codeword, denoted \( \mathbf{c}'_R \). Watermark recovery proceeds in two complementary ways: (i) each codeword in \( \mathbf{c}'_R \) is independently decoded via LDPC to produce candidate watermarks; or (ii) all codewords are aggregated via majority voting to form a consensus codeword \( \tilde{\mathbf{c}} \), which is then LDPC-decoded to reconstruct the original watermark \( \mathbf{w} \).

\subsection{Gaussian Shannon}

Existing methods tend to rely on fuzzy matching for watermark verification and traceability, which limits their application scenarios.This paper proposes a “Gaussian Shannon” watermarking method, which aims to ensure the integrity and robustness of watermark information. We have observed several phenomena: The process from sampling to performing DDIM inversion is analogous to the input-output process of a message transmission system. The input-output process transmits a bitstream representing the watermark information. The entire process is susceptible to noise interference, which generally comes from prediction errors of neural networks and adversarial attacks on images. Therefore, we can regard the process of embedding and extracting watermarks in diffusion models as a network communication process.Our goal is to reduce the impact of the noise in this process.

In terms of watermark extraction results, the communication outcome is primarily characterized by two types of errors. One is local errors, which is usually manifested as large - scale errors in the local positions of the latent space image, as shown in Fig.\ref{fig:fig4a}. Such errors can be compensated by other positions with fewer errors. The other type is the global random error, as shown in Fig.\ref{fig:fig4b}. This can be addressed using communication assurance measures.To address these two types of errors and ensure communication link reliability, we have developed a cascaded watermark embedding and recovery method that collaborates with majority voting and error-correction mechanisms.

\begin{figure}[h]
	\centering
	\setlength{\abovecaptionskip}{3pt} 
	\setlength{\belowcaptionskip}{-15pt} 
	\setlength{\tabcolsep}{2pt} 
	\begin{tabular}{@{}cccccc@{}}
		\includegraphics[width=0.15\linewidth]{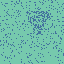}\customref{fig:fig4a}{4a} &
		\includegraphics[width=0.15\linewidth]{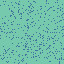}\customref{fig:fig4b}{4b} &
		\includegraphics[width=0.15\linewidth]{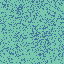}\customref{fig:fig4c}{4c} &
		\includegraphics[width=0.15\linewidth]{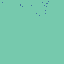}\customref{fig:fig4d}{4d} &
		\includegraphics[width=0.15\linewidth]{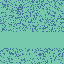}\customref{fig:fig4e}{4e} &
		\includegraphics[width=0.15\linewidth]{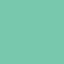}\customref{fig:fig4f}{4f} \\
		(a) & (b) & (c) & (d) & (e) & (f)
	\end{tabular}
	
	\caption{Error bits(dark) in the latent variable.
    (a) Local errors + global random errors.
		(b) Global random errors.
		(c) Errors at JPEG25.
		(d) Majority voting only for (c).
		(e) Error correction only for (c).
		(f) Using both methods for (c).}
\end{figure}

\subsection{Watermark Embedding}

Given the watermark sequence \(w=\{w_i\}_{i=1}^k\), where \(w_i\in\{0,1\}\), we first encode the watermark sequence using Low-Density Parity-Check (LDPC) codes:

\[c=\text{LDPC}_{\text{Encode}}(n,k,w)\]

Here, \(k\) and \(n\) are the information and codeword lengths, respectively, with the codeword \(c\in\{0,1\}^n\).

Let the total dimension of the latent space image be \(P = H\times W\times C\). In general, the encoded codeword length \(n\) is set to be a divisor of the dimension of the latent space image \(P\), so that the codeword can be redundantly expanded to enhance robustness. The codeword \(c\) is repeated \(R=P/n\) times to obtain the extended sequence \(c_{R}\). Next, the extended sequence is pseudo-randomized based on the key \(K\) to maintain a standard Gaussian distribution:
\[s=\text{Pseudo-Random}(K,c_{R})\in\{0,1\}^{P}\]

Finally, Gaussian sampling is performed to generate the watermarked initial noise. For each pixel position \(j\), we independently sample a standard Gaussian random variable \(\epsilon_j\sim\mathcal{N}(0,1)\) and compute its absolute value \(|\epsilon_j|\), which follows a half-normal distribution with parameter \(\sigma=1\). The initial noise with the watermark is defined as:
\[z_T^{j}=(-1)^{1-s_j}\cdot|\epsilon_j|,\quad j=1,2,...,P\]

The resulting noise tensor \(z_T\) is used to generate the final image through a gradual denoising process:
\[x=\text{DiffusionSampler}(z_T,\theta,T)\]

Here, \(\theta\) is the diffusion model parameter, and \(T\) is the total number of diffusion steps.

\subsection{Watermark Extraction}

The given image is processed through DDIMInversion to obtain the initial noise \(z_T\in\mathbb{R}^{H\times W\times C}\). Based on the pseudo-randomization key used in the embedding stage, the extended codeword \(c'_{R}\) can be parsed from the noise \(z_T\). The \(c'_{R}\) is then divided into \(R\) redundant codeword blocks of length \(n\): \(c'_{R}=[c_1,c_2,\dots,c_{r}]\) ,where \(c_r\in\{0,1\}^n\).

For each codeword \(c_r\), the LDPC error correction algorithm is used for decoding. If there exists a codeword \(c_r\) that satisfies the LDPC parity check equation \(H\cdot c_r^T=0 \pmod 2\) (\(H\) is the parity check matrix), then the information bit part \(w_r=[c_{r1},\dots,c_{rk}]\) is extracted as the final extracted watermark \(w'\). When none of the codeword \(c_r\) satisfy the parity check equation, a majority voting is performed. Specifically, a bit-by-bit voting is conducted on the values at the same bit positions of each codeword:
\[\tilde{c}_i=\text{mode}\{c_{1i},c_{2i},\dots,c_{ri}\},\quad i=1,2,\dots,n\]

This results in the composite codeword \(\tilde{c}\). The LDPC error correction is attempted again on \(\tilde{c}\). If the error correction is successful, the information bits are extracted as the watermark. Otherwise, the watermark information is considered to have a high bit error rate, and its integrity cannot be guaranteed. It is only used for verification:
\[
w'=
\begin{cases}
	\text{information bits}(c_r),&\exists r,H\cdot c_r^T=0 \pmod 2\\
	\text{information bits}(\tilde{c}),&H\cdot\tilde{c}^T=0 \pmod 2\\
	\text{only verification},&\text{else.}
\end{cases}
\]





\subsection{Robustness Analysis of Gaussian Shannon}

Considering the composite channel model faced by the watermarking system, where the diffusion generation, channel attacks, and reverse diffusion processes together form a binary input additive white gaussian noise channel(BIAWGN), a single error correction mechanism is difficult to ensure reliability in this complex environment. However, the cascaded scheme achieves synergistic gain through the complementarity of error correction.

\noindent\textbf{Capability of Majority Voting.} Majority voting, as a simple yet effective error - correcting mechanism, is based on the idea of exploiting spatial redundancy to enhance reliability. Consider a basic binary communication system in which a single bit is transmitted \( m \) times. At the receiver, we perform majority voting on the \( m \) received values: if more than half of the received values are 1, we decide that the original bit is 1; otherwise, it is 0. Let the probability of error in each transmission be \( p \), and assume that errors are independent of each other. The probability of error in majority voting is the probability that "more than half of the transmissions are erroneous." The formula is expressed as:

\[
P_{error}^{\text{maj}} = \sum_{k=\lceil m/2 \rceil}^{m} \binom{m}{k} p^k (1-p)^{m-k}
\]

where \( \binom{m}{k} \) is the binomial coefficient, representing the number of ways to choose \( k \) erroneous transmissions out of \( m \) transmissions.

Using the Chernoff bound from probability theory, we can obtain an upper bound on the error probability:

\[
P_{error}^{\text{maj}} \leq \exp\left(-m\cdot D\left(\frac{1}{2} \| p\right)\right)
\]

where \( D(a \| b) = a\ln\frac{a}{b} + (1-a)\ln\frac{1-a}{1-b} \).

When the error probability \(p < 0.5\), the exponential term is negative, and the error probability decays exponentially as \(m\) increases. This means that increasing the number of redundancy can quickly reduce the error rate. This property makes majority voting particularly suitable for solving local errors and improving the quality of the entire codeword, as shown in Fig.\ref{fig:fig4d}. However, when the error probability $p \geq 0.5$, the system fails. This precisely illustrates the necessity of introducing error - correcting codes.

\noindent\textbf{Random Error Correction Capability.} LDPC codes exhibit theoretical optimality in correcting random errors, and their error correction capability can be analyzed through their threshold characteristics. For a given LDPC code, there exists a critical threshold \(\text{SNR}^*\). When the actual SNR exceeds this threshold, the belief propagation algorithm is more likely to converge to the correct solution as the code length increases. The solution for this threshold can be expressed as a recursive equation:
\[
P_{l+1} = f(P_l, \text{SNR})
\]
where $P_l$ is the average bit error probability after the $l$-th iteration, SNR is the signal-to-noise ratio, and $f$ is the single-iteration error probability update rule.

The above process can be described as follows: starting from the initial channel conditions (i.e., different SNR ), iterative calculations are performed. When the SNR is high, \(P_l\) continuously decreases during iterations, indicating that errors are gradually corrected, and eventually, \(P_l\) tends to zero. When the SNR is low, \(P_l\) stagnates or tends to a non-zero value, indicating that the decoding is trapped in an error equilibrium. The bisection method allows us to identify the critical SNR for the system. This enables us to use limited code lengths and a limited number of iterations to make the probability of convergence to the correct solution fall within an acceptable range. In this paper, at an environmental SNR of \(0\text{ dB}\), the (3,4)-regular LDPC code with a code rate \(R=1/4\) exhibits good performance. This performance enables LDPC codes to effectively correct the random noise introduced in the diffusion process. However, its threshold property also indicates that in harsh channels, the channel conditions must meet the threshold requirements of LDPC codes in order to fully leverage their error - correcting capabilities,as shown in Fig.\ref{fig:fig4e}.

\noindent\textbf{Synergy of the Cascaded System.}The limitations of a single error - correcting mechanism prompt us to consider the combined performance of a two - level error - correcting system: majority voting improves the quality of the information, reducing the high error probability $p$ to the working range of LDPC codes. Subsequently, the LDPC codes perform the residual error correction. The two mechanisms complement each other, as shown in Fig.\ref{fig:fig4f}.

\section{Experiments}
\label{sec:formatting}
This section presents a thorough evaluation of the proposed method, which includes experimental results,
performance comparisons with state-of-the-art methods,
and ablation studies.

\begin{table*}[h]
	\centering
	\setlength{\abovecaptionskip}{3pt} 
	\setlength{\belowcaptionskip}{-15pt} 
	\adjustbox{max width=\textwidth}{%
		\begin{tabular}{l|cc|cc|cc}
			\toprule
			\multirow{2}{*}{Methods}&\multicolumn{2}{c|}{TPR@\(10^{-6}\)FPR} & \multicolumn{2}{c|}{BitAcc.} & \multicolumn{2}{c}{TPR@BitAcc.100\%.}\\
			\cmidrule(lr){2-7}
			&W/o Noise&W/Noise&W/o Noise&W/Noise&W/o Noise&W/Noise\\
		
			\cmidrule(lr){1-7}
			DwtDct\cite{cox2007digital}&0.836/0.890/0.887&0.165/0.174/0.178&0.8106/0.8153/0.8149&0.5683/0.5657/0.5652&0.047/0.029/0.036&0.020/0.016/0.014\\
			
			DwtDctSvd\cite{cox2007digital}&1.000/1.000/1.000&0.599/0.592/0.592&0.9996/0.9983/0.9982&0.7004/0.6884/0.6820&0.416/0.331/0.359&0.135/0.099/0.062\\
			
			Tree-Ring\cite{wen2023tree}&1.000/1.000/1.000&0.909/0.901/0.899&-&-&-&-\\
			
			StableSignature$_{\textrm{[ICCV2023]}}$\cite{fernandez2023stable}&1.000/1.000/1.000&0.679/0.654/0.675&0.9946/0.9932/0.9923&0.7642/0.7528/0.7540&0.736/0.725/0.734&0.173/0.179/0.182\\
			
			GaussianShading$_{\textrm{[CVPR2024]}}$\cite{yang2024gaussian}&1.000/1.000/1.000&0.999/0.999/0.999&0.9999/0.9999/0.9999&0.9716/0.9702/0.9691&0.989/0.989/0.988&0.399/0.387/0.381\\
			
			PRCW$_{\textrm{[ICLR2025]}}$\cite{gunn2024undetectable}&1.000/1.000/1.000&0.855/0.834/0.834&1.0000/1.0000/1.0000&0.9230/0.9156/0.9142&1.000/1.000/1.000&0.855/0.827/0.827\\
			
			\textbf{Ours}&\textbf{1.000/1.000/1.000}&\textbf{1.000/1.000/1.000}&\textbf{1.0000/1.0000/1.0000}&\textbf{0.9932/0.9926/0.9925}&\textbf{1.000/1.000/1.000}&\textbf{0.968/0.966/0.965}\\
			
			\bottomrule 
		\end{tabular}%
	}
	\caption{\textbf{Performance comparisons of our proposed method and previous state-of-the-art methods.}
	We evaluate TPR@\(10^{-6}\)FPR, BitAcc. and TPR@BitAcc.100\% on SDv1.4, 2.0, and 2.1, respectively. The “Noise” denotes the result under the average noise level.}
	\label{tab:tab1} 
\end{table*}

\subsection{Experimental Setup}


\textbf{Baselines.} We comprehensively compare our proposed method with 6 baseline method, including: DwtDct \cite{cox2007digital}, DwtDctSvd \cite{cox2007digital}, Stable Signature \cite{fernandez2023stable}, Tree-Ring \cite{wen2023tree}, and performance lossless state-of-the-art methods such as Gaussian Shading \cite{yang2024gaussian} and PRCW \cite{gunn2024undetectable}. 

\noindent\textbf{Implementation Details.} To evaluate the performance of our proposed Gaussian Shannon, we follow baseline methods to use three versions of stable diffusion:  V1.4, V2.0, and V2.1. The size of the generated images is \(512 \times 512\), with a latent space dimension of \(4 \times 64 \times 64\). During inference, we use prompts from the prompt dataset on Hugging Face, with a guidance scale of 7.5. For ODE-based samplers, we use DDIM\cite{song2020denoising} for 50-step sampling. 
Since watermark detectors are usually unaware of the prompt used to generate images, we use an empty prompt with a guidance scale of 1 for inversion, employing DDIM inversion~\cite{song2020denoising} for 50 steps.
Each experiment generates 1000 images. To facilitate comparison with other methods, we fix the watermark capacity at 256 bits. Other parameters are set by default as redundancy \(m = 16\), code rate \(R = 0.25\), and channel signal-to-noise ratio (SNR) at 0 dB. All experiments are conducted using the PyTorch 2.5.1
framework, running on a single RTX 4090 GPU.

\begin{figure}[h]
	\centering
	\setlength{\abovecaptionskip}{3pt} 
	\setlength{\belowcaptionskip}{-10pt} 
	\setlength{\tabcolsep}{2pt} 
	\begin{tabular}{@{}ccccc@{}}
		\includegraphics[width=0.19\linewidth]{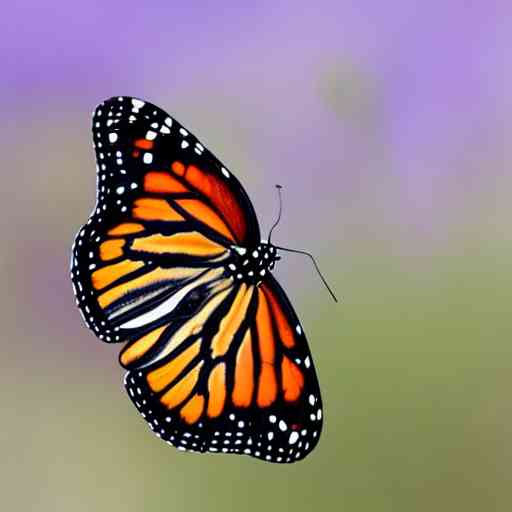}\customref{fig:fig5a}{5a} &
		\includegraphics[width=0.19\linewidth]{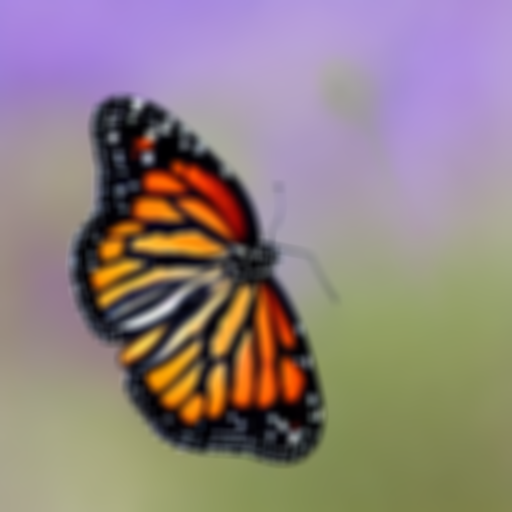}\customref{fig:fig5b}{5b} &
		\includegraphics[width=0.19\linewidth]{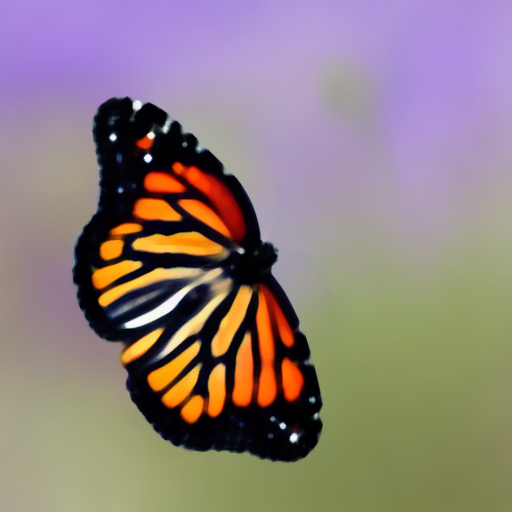}\customref{fig:fig5c}{5c} &
		\includegraphics[width=0.19\linewidth]{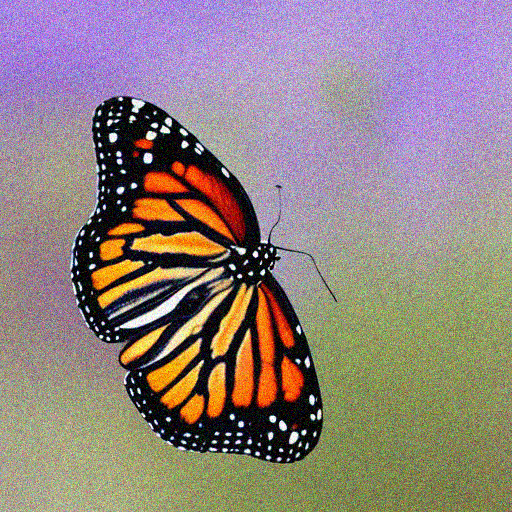}\customref{fig:fig5d}{5d} &
		\includegraphics[width=0.19\linewidth]{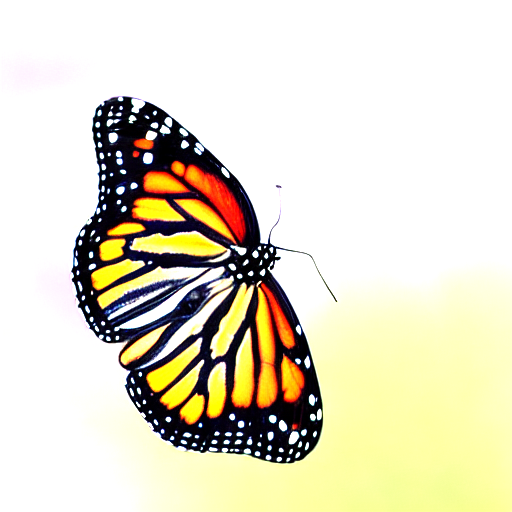}\customref{fig:fig5e}{5e} \\
		(a) & (b) & (c) & (d) & (e) \\
		\includegraphics[width=0.19\linewidth]{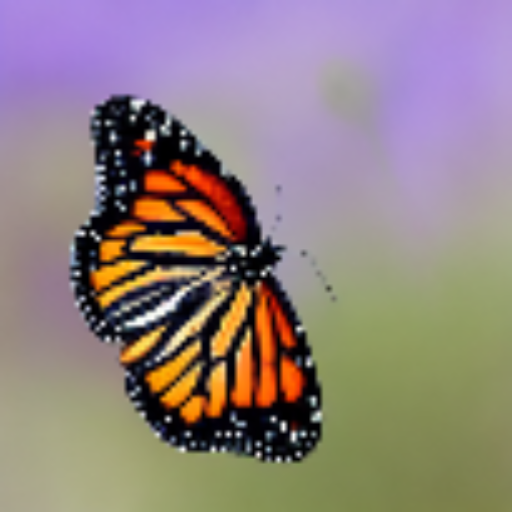}\customref{fig:fig5f}{5f} &
		\includegraphics[width=0.19\linewidth]{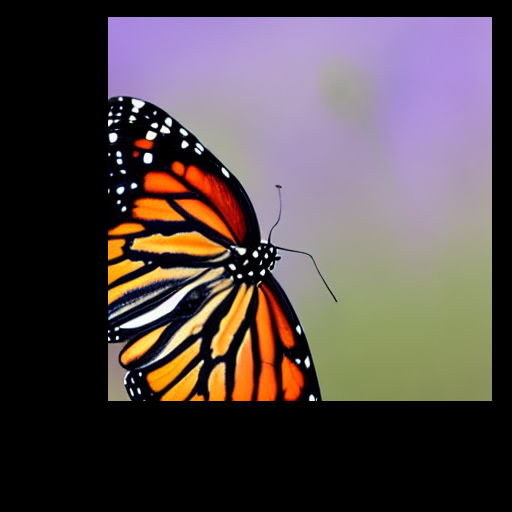}\customref{fig:fi5g}{5g} &
		\includegraphics[width=0.19\linewidth]{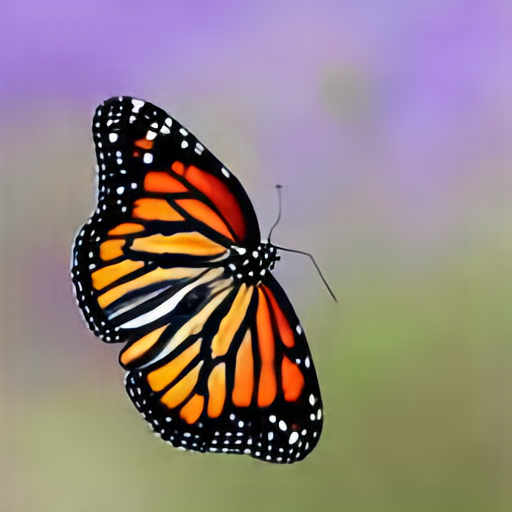}\customref{fig:fig5h}{5h} &
		\includegraphics[width=0.19\linewidth]{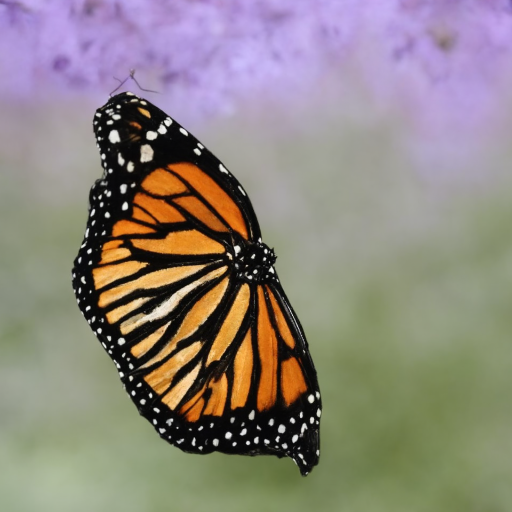}\customref{fig:fig5i}{5i} &
		\includegraphics[width=0.19\linewidth]{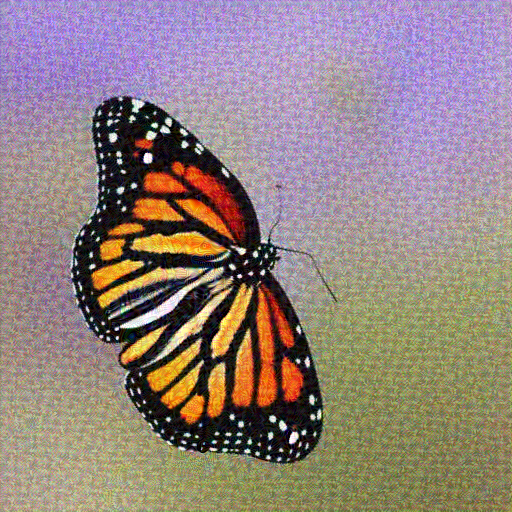}\customref{fig:fig5j}{5j} \\
		(f) & (g) & (h) & (i) & (j) 
	\end{tabular}
	\caption{Watermarked images under different noises or attacks. (a) JPEG quality factor 25.
 (b) Gaussian blur radius=4. (c) Median filter k=7. (d) Gaussian noise $\sigma$ =0.05
 (e) Brightness factor 2. (f) Scaling 0.3. (g) Random drop 0.25. (h) VAE compression (i) Diffusion attack. (j) Embedding attack.}
 \label{fig:noise}
\end{figure}

\noindent\textbf{Robustness Evaluation.} The robustness of our method is evaluated under seven representative noise conditions, detailed in Fig.~\ref{fig:noise}. For each noise type, we applied the strength specified in the figure.

\noindent\textbf{Evaluation Metrics.} 
We calculate the average bit accuracy (BitAcc.) and the true positive rate corresponding to the bit threshold at a fixed false positive rate (TPR@\(10^{-6}\)FPR). To emphasize information integrity, we calculate the true positive rate (TPR) at 100\% bit accuracy (TPR@BitAcc.100\%) as the primary evaluation metric, and the majority voting rate as the method performance metric. Furthermore, we utilize FID and CLIP Score to evaluate image quality.


\subsection{Performance of Gaussian Shannon}

\textbf{Conventional Detection and Tracing.} We first calculate the threshold $\tau = 65 +$ corresponding to a fixed false positive rate (FPR) less than $10^{-6}$, and then compute the true positive rate (TPR) on watermarked images, considering that exceeding this threshold is detectable. Performance tests are conducted in seven specific noise environments of varying intensity, as shown in Fig.\ref{fig:fig6}. Our method maintains a 99\% TPR at a threshold of 75\% and exhibits robust performance. In addition, we conducted experiments with varying noise intensities (Fig.\ref{fig:fig7}) to determine the performance limit of our method.

In the traceability scenario, we assign a unique watermark to each user. The extracted watermark is matched with the user ID, and the image owner is identified as the individual whose bit hits exceed the threshold $\tau \ge 65$ and who has the highest number of bit matches. As shown in Fig.\ref{fig:fig6}, our method maintains a high traceability rate under various noise conditions.

\noindent\textbf{Information Parsing Scenario.} Our method advocates information integrity, using TPR@BitAcc.100\% as the metric. Performance experiments are conducted in the same seven specific noise environments of varying intensity, as shown in Fig.\ref{fig:fig6}, with the threshold $\tau = 100$. An accuracy rate of over 92\% is achieved.

\begin{figure}[h] 
	\centering 
	\setlength{\abovecaptionskip}{2pt} 
	\begin{minipage}{0.49\linewidth}
	\resizebox{1\columnwidth}{!}{ 
		\begin{tikzpicture} 
			\begin{axis}[
				sharp plot, 
				title=,
				xmode=normal,
				ymode=normal,
				xlabel=threshold $\tau$, 
				ylabel=True Positive Rate, 
				width=8cm, height=6cm,  
				xmin=0.63,xmax=1.02,  
				ymin=0.90,ymax=1.01,  
				xtick={0.65,0.70,0.75,0.80,0.85,0.9,0.95,1.00}, 
				ytick={0.92,0.94,0.96,0.98,1.00}, 
				xticklabel={$\pgfmathprintnumber[fixed, fixed zerofill, precision=2]{\tick}$},
				yticklabel={$\pgfmathprintnumber[fixed, fixed zerofill, precision=2]{\tick}$},
				xlabel style={font=\footnotesize},
				ylabel style={font=\footnotesize},
				ticklabel style={font=\scriptsize},  
				legend style={
					at={(0.4,0.6)},
					font=\scriptsize,  
					draw=none,
				},
				xlabel near ticks, 
				ylabel near ticks, 
				xmajorgrids=false, 
				ymajorgrids=false, 
				grid style=dashed, 
				legend  style={
					at={(0.4,0.65)},
					font=\fontsize{6}{1}\selectfont,
					draw=none, 
				},
				tick align=outside,  
				xtick pos=bottom,    
				ytick pos=left,      
				]
				
				\addplot+[solid, thick,mark options={scale=0.5}, color=color0] 
				plot coordinates { 
					(0.65,1)
					(0.675,1)
					(0.70,1)
					(0.725,1)
					(0.75,1)
					(0.775,1)
					(0.80,1)
					(0.825,1)
					(0.85,1)
					(0.875,1)
					(0.90,1)
					(0.925,1)
					(0.95,1)
					(0.975,1)
					(1,1)
				};
				\addlegendentry{None}
				
				\addplot+[solid, thick,mark options={scale=0.5}, color=color2] 
				plot coordinates {
					(0.65,1)
					(0.675,1)
					(0.70,0.998)
					(0.725,0.997)
					(0.75,0.996)
					(0.775,0.993)
					(0.80,0.991)
					(0.825,0.986)
					(0.85,0.982)
					(0.875,0.9815)
					(0.90,0.981)
					(0.925,0.980)
					(0.95,0.979)
					(0.975,0.979)
					(1,0.979)
				};
				\addlegendentry{Gaussian Blur} 
				
				\addplot+[solid, thick,mark options={scale=0.5}, color=color3] 
				plot coordinates {

					(0.65,0.999)
					(0.675,0.999)
					(0.70,0.998)
					(0.725,0.996)
					(0.75,0.995)
					(0.775,0.9945)
					(0.80,0.994)
					(0.825,0.992)
					(0.85,0.989)
					(0.875,0.988)
					(0.90,0.987)
					(0.925,0.9862)
					(0.95,0.986)
					(0.975,0.986)
					(1,0.986)
				};
				\addlegendentry{Median Filter} 
				
				\addplot+[solid, thick,mark options={scale=0.5}, color=color4] 
				plot coordinates {

					(0.65,1)
					(0.675,1)
					(0.70,1)
					(0.725,1)
					(0.75,0.998)
					(0.775,0.992)
					(0.80,0.989)
					(0.825,0.985)
					(0.85,0.979)
					(0.875,0.976)
					(0.90,0.972)
					(0.925,0.971)
					(0.95,0.970)
					(0.975,0.970)
					(1,0.970)
				};
				\addlegendentry{Gaussian Noise} 
				
				\addplot+[solid, thick,mark options={scale=0.5}, color=color5] 
				plot coordinates {

					(0.65,1)
					(0.675,1)
					(0.70,1)
					(0.725,1)
					(0.75,0.999)
					(0.775,0.997)
					(0.80,0.995)
					(0.825,0.992)
					(0.85,0.990)
					(0.875,0.987)
					(0.90,0.986)
					(0.925,0.983)
					(0.95,0.982)
					(0.975,0.982)
					(1,0.982)
				};
				\addlegendentry{Brightness} 
				
				\addplot+[solid, thick,mark options={scale=0.5}, color=color6] 
				plot coordinates {
			
					(0.65,1)
					(0.675,0.996)
					(0.70,0.993)
					(0.725,0.991)
					(0.75,0.989)
					(0.775,0.980)
					(0.80,0.971)
					(0.825,0.962)
					(0.85,0.953)
					(0.875,0.949)
					(0.90,0.942)
					(0.925,0.940)
					(0.95,0.938)
					(0.975,0.938)
					(1,0.938)
				};
				\addlegendentry{jpeg} 
				
				\addplot+[solid, thick,mark options={scale=0.5}, color=color7] 
				plot coordinates {
				
					(0.65,1)
					(0.675,1)
					(0.70,1)
					(0.725,1)
					(0.75,1)
					(0.775,0.993)
					(0.80,0.986)
					(0.825,0.979)
					(0.85,0.971)
					(0.875,0.963)
					(0.90,0.956)
					(0.925,0.955)
					(0.95,0.953)
					(0.975,0.953)
					(1,0.953)
				};
				\addlegendentry{Resize} 
				
				\addplot+[solid,thick,mark options={scale=0.5}, color=color1] 
				plot coordinates {
			
					(0.65,1)
					(0.675,1)
					(0.70,1)
					(0.725,1)
					(0.75,1)
					(0.775,0.996)
					(0.80,0.992)
					(0.825,0.991)
					(0.85,0.990)
					(0.875,0.985)
					(0.90,0.977)
					(0.925,0.975)
					(0.95,0.972)
					(0.975,0.972)
					(1,0.972)
				};
				\addlegendentry{Random Drop} 
				
			\end{axis}
		\end{tikzpicture}
	}
	\caption*{(a) Detection results.} 
	\end{minipage}
	\hfill
		\begin{minipage}{0.49\linewidth}
	\resizebox{1\columnwidth}{!}{ 
		\begin{tikzpicture} 
			\begin{axis}[
				sharp plot, 
				title=,
				xmode=log,
				ymode=normal,
				xlabel=Number of Users, 
				ylabel=Accuracy of Identification, 
				width=8cm, height=6cm,  
				xmin=5,xmax=15000000,  
				ymin=0.955,ymax=1.005,  
				xtick={1e1, 1e2, 1e3, 1e4, 1e5, 1e6, 1e7},
				ytick={0.96,0.97,0.98,0.99,1.00}, 
				every minor tick/.style={draw=none},  
				yticklabel={$\pgfmathprintnumber[fixed, fixed zerofill, precision=2]{\tick}$},
				xlabel style={font=\footnotesize},
				ylabel style={font=\footnotesize},
				ticklabel style={font=\scriptsize},  
				legend style={
					at={(0.4,0.6)},
					font=\scriptsize,  
					draw=none,
				},
				xlabel near ticks, 
				ylabel near ticks, 
				xmajorgrids=false, 
				ymajorgrids=false, 
				grid style=dashed, 
				legend  style={
					at={(0.4,0.65)},
					font=\fontsize{6}{1}\selectfont,
					draw=none, 
				},
				tick align=outside,  
				xtick pos=bottom,    
				ytick pos=left,      
				]
				
				\addplot+[solid, thick,mark options={scale=0.5}, color=color0] 
				plot coordinates { 
					(1e1,1)
					(1e2,1)
					(1e3,1)
					(1e4,1)
					(1e5,1)
					(1e6,1)
					(1e7,1)
				};
				\addlegendentry{None}
			
				\addplot+[solid, thick,mark options={scale=0.5}, color=color2] 
				plot coordinates {		
					(1e1,1)
					(1e2,0.999)
					(1e3,0.998)
					(1e4,0.9975)
					(1e5,0.9973)
					(1e6,0.9972)
					(1e7,0.9962)
				};
				\addlegendentry{Gaussian Blur} 
				
				\addplot+[solid, thick,mark options={scale=0.5}, color=color3] 
				plot coordinates {
					(1e1,1)
					(1e2,0.999)
					(1e3,0.9985)
					(1e4,0.9979)
					(1e5,0.9972)
					(1e6,0.9968)
					(1e7,0.9964)
				};
				\addlegendentry{Median Filter} 
				
				\addplot+[solid, thick,mark options={scale=0.5}, color=color4] 
				plot coordinates {
					(1e1,1)
					(1e2,0.9995)
					(1e3,0.9986)
					(1e4,0.9965)
					(1e5,0.9942)
					(1e6,0.9923)
					(1e7,0.9895)
				};
				\addlegendentry{Gaussian Noise} 
				
				\addplot+[solid, thick,mark options={scale=0.5}, color=color5] 
				plot coordinates {
					(1e1,1)
					(1e2,0.9982)
					(1e3,0.9954)
					(1e4,0.9936)
					(1e5,0.9924)
					(1e6,0.9915)
					(1e7,0.9891)
				};
				\addlegendentry{Brightness} 
				
				\addplot+[solid, thick,mark options={scale=0.5}, color=color6] 
				plot coordinates {
					(1e1,1)
					(1e2,0.999)
					(1e3,0.9975)
					(1e4,0.9973)
					(1e5,0.9967)
					(1e6,0.9954)
					(1e7,0.9936)
				};
				\addlegendentry{jpeg} 
				
				\addplot+[solid, thick,mark options={scale=0.5}, color=color7] 
				plot coordinates {
					(1e1,1)
					(1e2,0.999)
					(1e3,0.998)
					(1e4,0.9972)
					(1e5,0.9971)
					(1e6,0.9968)
					(1e7,0.9953)
				};
				\addlegendentry{Resize} 
				
				\addplot+[solid,thick,mark options={scale=0.5}, color=color1] 
				plot coordinates {
					(1e1,1)
					(1e2,0.995)
					(1e3,0.992)
					(1e4,0.9898)
					(1e5,0.9884)
					(1e6,0.9873)
					(1e7,0.9864)
				};
				\addlegendentry{Random Drop} 
				
			\end{axis}
		\end{tikzpicture}
		}
		\caption*{(b) Traceability results.} 
	\end{minipage}\\
	\setlength{\abovecaptionskip}{3pt} 
	\setlength{\belowcaptionskip}{-8pt} 
	\caption{Performance comparisons of our proposed method on different noises. (a): The x-axis represents the threshold \(\tau\) at different fixed FPR, and the y-axis indicates TPR. (b): The x-axis represents the number of users accommodated by the watermark, and the y-axis represents the traceability rate. 
	} 
	\label{fig:fig6}

\end{figure}	

\begin{figure*}[h] 
	
	\centering 
	\setlength{\belowcaptionskip}{-3pt} 
	\begin{minipage}{0.24\textwidth}  
		
		\resizebox{1\columnwidth}{!}{ 
			\begin{tikzpicture} 
				\begin{axis}[
					sharp plot, 
					title=,
					xmode=normal,
					ymode=normal,
					xlabel=JPEG Quality Factor, 
					ylabel=Bit Accuracy/True Positive Rate, 
					width=8cm, height=6cm,  
					x dir=reverse,
					xmin=5,xmax=95,  
					ymin=0.0,ymax=1.05,  
					xtick={90,70,50,30,10},  
					ytick={0,0.2,0.4,0.6,0.8,1.00}, 
					xlabel near ticks, 
					ylabel near ticks, 
					xmajorgrids=false, 
					ymajorgrids=false, 
					grid style=dashed, 
					legend style={
						at={(0.55,0.4)},
						font=\fontsize{8}{1}\selectfont,
						draw=none, 
					},
					tick align=outside,  
					xtick pos=bottom,    
					ytick pos=left,      
					]
					
					\addplot+[solid,very thick,mark=diamond*,mark options={scale=0.6}, color=color1] 
					plot coordinates { 
						(90,1)
						(70,0.995)
						(50,0.976)
						(30,0.938)
						(10,0.672)
					};
					\addlegendentry{TPR@BitAcc.100\%}
					
					\addplot+[solid,very thick,mark=*,mark options={scale=0.6}, color=color2] 
					plot coordinates {
						(90,1)
						(70,1)
						(50,1)
						(30,0.997)
						(10,0.987)
					};
					\addlegendentry{TPR@10-6FPR} 
					
					\addplot+[solid,very thick,mark=square*,mark options={scale=0.6}, color=color3] 
					plot coordinates {
						(90,1)
						(70,0.998)
						(50,0.994)
						(30,0.983)
						(10,0.913)
					};
					\addlegendentry{BitAcc.} 
					
				\end{axis}
			\end{tikzpicture}
		}
		\caption*{(a) jpeg}  
	\end{minipage}
	\hfill
	\begin{minipage}{0.24\textwidth}  
		
		\resizebox{1\columnwidth}{!}{ 
			\begin{tikzpicture} 
				\begin{axis}[
					sharp plot, 
					title=,
					xmode=normal,
					ymode=normal,
					xlabel=Convolution Kernel Radius, 
					ylabel=Bit Accuracy/True Positive Rate, 
					width=8cm, height=6cm,  
					xmin=1.5,xmax=8.5,  
					ymin=0,ymax=1.05,  
					xtick={2,3,4,5,6,7,8}, 
					ytick={0,0.2,0.4,0.6,0.8,1.00}, 
					xlabel near ticks, 
					ylabel near ticks, 
					xmajorgrids=false, 
					ymajorgrids=false, 
					grid style=dashed, 
					legend style={
						at={(0.55,0.4)},
						font=\fontsize{8}{1}\selectfont,
						draw=none, 
					},
					tick align=outside,  
					xtick pos=bottom,    
					ytick pos=left,      
					]
					
					\addplot+[solid,very thick,mark=diamond*,mark options={scale=0.6}, color=color1] 
					plot coordinates { 
						(2,0.994)
						(4,0.979)
						(6,0.576)
						(8,0.005)
					};
					\addlegendentry{TPR@BitAcc.100\%}
					
					\addplot+[solid,very thick,mark=*,mark options={scale=0.6}, color=color2] 
					plot coordinates {
						(2,1)
						(4,1)
						(6,0.997)
						(8,0.791)
					};
					\addlegendentry{TPR@10-6FPR} 
					
					\addplot+[solid,very thick,mark=square*,mark options={scale=0.6}, color=color3] 
					plot coordinates {
						(2,0.998)
						(4,0.995)
						(6,0.902)
						(8,0.696)
					};
					\addlegendentry{BitAcc.} 
					
				\end{axis}
			\end{tikzpicture}
		}
		\caption*{(b) Gaussian Blur}  
	\end{minipage}
	\hfill
	\begin{minipage}{0.24\textwidth}  
		
		\resizebox{1\columnwidth}{!}{ 
			\begin{tikzpicture} 
				\begin{axis}[
					sharp plot, 
					title=,
					xmode=normal,
					ymode=normal,
					xlabel=Filter Kernel Size, 
					ylabel=Bit Accuracy/True Positive Rate, 
					width=8cm, height=6cm,  
					xmin=2,xmax=20,  
					ymin=0.0,ymax=1.05,  
					xtick={3,7,11,15,19}, 
					ytick={0,0.2,0.4,0.6,0.8,1.00}, 
					xlabel near ticks, 
					ylabel near ticks, 
					xmajorgrids=false, 
					ymajorgrids=false, 
					grid style=dashed, 
					legend style={
						at={(0.55,0.4)},
						font=\fontsize{8}{1}\selectfont,
						draw=none, 
					},
					tick align=outside,  
					xtick pos=bottom,    
					ytick pos=left,      
					]
					
					\addplot+[solid,very thick,mark=diamond*,mark options={scale=0.6}, color=color1] 
					plot coordinates { 
						(3,0.995)
						(7,0.986)
						(11,0.833)
						(15,0.102)
						(19,0.004)
					};
					\addlegendentry{TPR@BitAcc.100\%}
					
					\addplot+[solid,very thick,mark=*,mark options={scale=0.6}, color=color2] 
					plot coordinates {
						(3,1)
						(7,0.999)
						(11,0.999)
						(15,0.943)
						(19,0.755)
					};
					\addlegendentry{TPR@10-6FPR} 
					
					\addplot+[solid,very thick,mark=square*,mark options={scale=0.6}, color=color3] 
					plot coordinates {
						(3,0.998)
						(7,0.997)
						(11,0.970)
						(15,0.780)
						(19,0.696)
					};
					\addlegendentry{BitAcc.} 
					
				\end{axis}
			\end{tikzpicture}
		}
		\caption*{(c) Median Filter}  
	\end{minipage}
	\begin{minipage}{0.24\textwidth}  
		
		\resizebox{1\columnwidth}{!}{ 
			\begin{tikzpicture} 
				\begin{axis}[
					sharp plot, 
					title=,
					xmode=normal,
					ymode=normal,
					xlabel=Noise Standard Deviation, 
					ylabel=Bit Accuracy/True Positive Rate, 
					width=8cm, height=6cm,  
					xmin=0.0,xmax=0.42,  
					ymin=0.0,ymax=1.05,  
					scaled ticks=false,
					xticklabel style={/pgf/number format/fixed},
					xtick={0,0.05,0.1,0.15,0.2,0.25,0.3,0.35,0.4}, 
					ytick={0,0.2,0.4,0.6,0.8,1.00}, 
					xlabel near ticks, 
					ylabel near ticks, 
					xmajorgrids=false, 
					ymajorgrids=false, 
					grid style=dashed, 
					legend style={
						at={(0.55,0.4)},
						font=\fontsize{8}{1}\selectfont,
						draw=none, 
					},
					tick align=outside,  
					xtick pos=bottom,    
					ytick pos=left,      
					]
					
					\addplot+[solid,very thick,mark=diamond*,mark options={scale=0.6}, color=color1] 
					plot coordinates { 
						(0.05,0.970)
						(0.07,0.938)
						(0.1,0.872)
						(0.2,0.634)
						(0.4,0.224)
					};
					\addlegendentry{TPR@BitAcc.100\%}
					
					\addplot+[solid,very thick,mark=*,mark options={scale=0.6}, color=color2] 
					plot coordinates {
						(0.05,1)
						(0.07,0.999)
						(0.1,0.999)
						(0.2,0.984)
						(0.4,0.934)
					};
					\addlegendentry{TPR@10-6FPR} 
					
					\addplot+[solid,very thick,mark=square*,mark options={scale=0.6}, color=color3] 
					plot coordinates {
						(0.05,0.994)
						(0.07,0.986)
						(0.1,0.971)
						(0.2,0.909)
						(0.4,0.798)
					};
					\addlegendentry{BitAcc.} 
					
				\end{axis}
			\end{tikzpicture}
		}
		\caption*{(d) Gaussian Noise}  
	\end{minipage}
	\begin{minipage}{0.24\textwidth}  
		
		\resizebox{1\columnwidth}{!}{ 
			\begin{tikzpicture} 
				\begin{axis}[
					sharp plot, 
					title=,
					xmode=normal,
					ymode=normal,
					xlabel=Brightness Factor, 
					ylabel=Bit Accuracy/True Positive Rate, 
					width=8cm, height=6cm,  
					xmin=1.5,xmax=10.5,  
					ymin=0.0,ymax=1.05,  
					xtick={2,4,6,8,10},  
					ytick={0,0.2,0.4,0.6,0.8,1.00}, 
					xlabel near ticks, 
					ylabel near ticks, 
					xmajorgrids=false, 
					ymajorgrids=false, 
					grid style=dashed, 
					legend style={
						at={(0.55,0.4)},
						font=\fontsize{8}{1}\selectfont,
						draw=none, 
					},
					tick align=outside,  
					xtick pos=bottom,    
					ytick pos=left,      
					]
					
					\addplot+[solid,very thick,mark=diamond*,mark options={scale=0.6}, color=color1] 
					plot coordinates { 
						(2,0.981)
						(4,0.741)
						(6,0.530)
						(8,0.272)
						(10,0)
					};
					\addlegendentry{TPR@BitAcc.100\%}
					
					\addplot+[solid,very thick,mark=*,mark options={scale=0.6}, color=color2] 
					plot coordinates {
						(2,1)
						(4,0.983)
						(6,0.939)
						(8,0.901)
						(10,0.836)
					};
					\addlegendentry{TPR@10-6FPR} 
					
					\addplot+[solid,very thick,mark=square*,mark options={scale=0.6}, color=color3] 
					plot coordinates {
						(2,0.994)
						(4,0.926)
						(6,0.858)
						(8,0.823)
						(10,0.801)
					};
					\addlegendentry{BitAcc.} 
					
				\end{axis}
			\end{tikzpicture}
		}
		\caption*{(e) Brightness}  
	\end{minipage}
	\hfill
	\begin{minipage}{0.24\textwidth}  
		
		\resizebox{1\columnwidth}{!}{ 
			\begin{tikzpicture} 
				\begin{axis}[
					sharp plot, 
					title=,
					xmode=normal,
					ymode=normal,
					xlabel=Random Scaling Ratio, 
					ylabel=Bit Accuracy/True Positive Rate, 
					width=8cm, height=6cm,  
					x dir=reverse,
					xmin=0.05,xmax=0.95,  
					ymin=0.0,ymax=1.05,  
					xtick={0.1,0.3,0.5,0.7,0.9}, 
					ytick={0,0.2,0.4,0.6,0.8,1.00}, 
					xlabel near ticks, 
					ylabel near ticks, 
					xmajorgrids=false, 
					ymajorgrids=false, 
					grid style=dashed, 
					legend style={
						at={(0.55,0.4)},
						font=\fontsize{8}{1}\selectfont,
						draw=none, 
					},
					tick align=outside,  
					xtick pos=bottom,    
					ytick pos=left,      
					]
					
					\addplot+[solid,very thick,mark=diamond*,mark options={scale=0.6}, color=color1] 
					plot coordinates { 
						(0.9,1)
						(0.7,1)
						(0.5,1)
						(0.3,0.946)
						(0.1,0.02)
					};
					\addlegendentry{TPR@BitAcc.100\%}
					
					\addplot+[solid,very thick,mark=*,mark options={scale=0.6}, color=color2] 
					plot coordinates {
						(0.9,1)
						(0.7,1)
						(0.5,1)
						(0.3,1)
						(0.1,0.899)
					};
					\addlegendentry{TPR@10-6FPR} 
					
					\addplot+[solid,very thick,mark=square*,mark options={scale=0.6}, color=color3] 
					plot coordinates {
						(0.9,1)
						(0.7,1)
						(0.5,0.999)
						(0.3,0.987)
						(0.1,0.717)
					};
					\addlegendentry{BitAcc.} 
					
				\end{axis}
			\end{tikzpicture}
		}
		\caption*{(f) Resize}  
	\end{minipage}
	\hfill
	\begin{minipage}{0.24\textwidth}  
		
		\resizebox{1\columnwidth}{!}{ 
			\begin{tikzpicture} 
				\begin{axis}[
					sharp plot, 
					title=,
					xmode=normal,
					ymode=normal,
					xlabel=Random Drop Ratio, 
					ylabel=Bit Accuracy/True Positive Rate, 
					width=8cm, height=6cm,  
					xmin=0,xmax=0.55,  
					ymin=0.00,ymax=1.05,  
					xtick={0,0.1,0.2,0.3,0.4,0.5}, 
					ytick={0,0.2,0.4,0.6,0.8,1.00}, 
					xlabel near ticks, 
					ylabel near ticks, 
					xmajorgrids=false, 
					ymajorgrids=false, 
					grid style=dashed, 
					legend style={
						at={(0.55,0.4)},
						font=\fontsize{8}{1}\selectfont,
						draw=none, 
					},
					tick align=outside,  
					xtick pos=bottom,    
					ytick pos=left,      
					]
					
					\addplot+[solid,very thick,mark=diamond*,mark options={scale=0.6}, color=color1] 
					plot coordinates { 
						(0.05,1)
						(0.1,1)
						(0.2,1)
						(0.25,0.972)
						(0.3,0.720)
						(0.4,0.453)
						(0.5,0.0)
					};
					\addlegendentry{TPR@BitAcc.100\%}
					
					\addplot+[solid,very thick,mark=*,mark options={scale=0.6}, color=color2] 
					plot coordinates {
						(0.05,1)
						(0.1,1)
						(0.2,1)
						(0.25,1)
						(0.3,1)
						(0.4,0.988)
						(0.5,0.976)
					};
					\addlegendentry{TPR@10-6FPR} 
					
					\addplot+[solid,very thick,mark=square*,mark options={scale=0.6}, color=color3] 
					plot coordinates {
						(0.05,1)
						(0.1,1)
						(0.2,0.993)
						(0.25,0.979)
						(0.3,0.965)
						(0.4,0.862)
						(0.5,0.720)
					};
					\addlegendentry{BitAcc.} 
					
				\end{axis}
			\end{tikzpicture}
		}
		\caption*{(g) Random Drop}  
	\end{minipage}
	\hfill
	\begin{minipage}{0.24\textwidth}  
		\resizebox{1\columnwidth}{!}{ 
			\begin{tikzpicture} 
				\begin{axis}[
					sharp plot, 
					title=,
					xmode=normal,
					ymode=normal,
					xlabel=CFG, 
					ylabel=TPR/BitAcc./Vote Rate, 
					width=8cm, height=6cm,  
					xmin=0,xmax=18.5,  
					ymin=0.0,ymax=1.05,  
					xtick={0,2,4,6,8,10,12,14,16,18}, 
					ytick={0,0.2,0.4,0.6,0.8,1.00}, 
					xlabel near ticks, 
					ylabel near ticks, 
					xmajorgrids=false, 
					ymajorgrids=false, 
					grid style=dashed, 
					legend style={
						at={(0.55,0.5)},
						font=\fontsize{8}{1}\selectfont,
						draw=none, 
					},
					tick align=outside,  
					xtick pos=bottom,    
					ytick pos=left,      
					]
					
					\addplot+[solid,very thick,mark=diamond*,mark options={scale=0.6}, color=color1] 
					plot coordinates { 
						(2,1)
						(5,1)
						(7.5,1)
						(10,1)
						(14,1)
						(18,1)
					};
					\addlegendentry{TPR@BitAcc.100\%}
					
					\addplot+[dashed,thin,mark=*,mark options={scale=0.6}, color=color2] 
					plot coordinates {
						(2,1)
						(5,1)
						(7.5,1)
						(10,1)
						(14,1)
						(18,1)
					}; 
					\addlegendentry{TPR@10-6FPR}
					
					\addplot+[dotted,thick,mark=square*,mark options={scale=0.6}, color=color3] 
					plot coordinates {
						(2,1)
						(5,1)
						(7.5,1)
						(10,1)
						(14,1)
						(18,1)
					}; 
					\addlegendentry{BitAcc.}
					
					\addplot+[solid,very thick,mark=triangle*,mark options={scale=0.6}, color=color4] 
					plot coordinates {
						(2,0.001)
						(6,0.011)
						(7.5,0.028)
						(10,0.041)
						(14,0.091)
						(18,0.142)
					}; 
					\addlegendentry{Vote Rate} 
					
				\end{axis}
			\end{tikzpicture}
		}
		\caption*{(i) Guidance scale}  
		\customref{fig:fig7i}{7i} 
	\end{minipage}
	\setlength{\belowcaptionskip}{-15pt} 
	\caption{
		Experimental results under different intensities of 7 types of noise, with the last one being the results under different guidance scales. The x-axis represents the intensity of the variable. The different curves correspond to the metrics: TPR@\(10^{-6}\)FPR, BitAcc. and TPR@BitAcc.100\%. The y-axis shows the results for these respective metrics.
	} 
	\label{fig:fig7}
\end{figure*}

\noindent\textbf{Image Quality.} We evaluate the differences in the quality of generated images using CLIP - Score and FID metrics. As shown in the Tab.\ref{tab:tab2}, there is hardly any difference in the quality of generated images between our method and other semantic watermarking methods.


\noindent\textbf{Sampler.} We conduct experiments on  various ODE - based sampling methods\cite{song2020denoising,lu2022dpm,zhao2023unipc,liu2022pseudo,zhang2022fast}. As shown in Tab.\ref{tab:tab3}, when considering strict conditions such as information integrity, there are some performance differences for our method across different samplers. Essentially, it is still a matter of whether the inversion can be accurately performed. It is just that the strict criteria have magnified the originally minor differences.

\begin{table}[h]
	
	\centering
	\setlength{\abovecaptionskip}{5pt}  
	\setlength{\belowcaptionskip}{-10pt} 
	\adjustbox{max width=1\linewidth}{%
		
		\begin{tabular}{l|c|c}
			\toprule
			
			Methods &CLIP Score ↑&FID↓ \\
			
			\midrule
			
			StableDiffusion&\textbf{0.3576/0.3611/0.3615}&\textbf{24.32}/24.40/\textbf{24.26}\\
			
			DwtDct\cite{cox2007digital}&0.3526/0.3598/0.3591&25.01/24.93/24.98\\
			
			DwtDctSvd\cite{cox2007digital}&0.3532/0.3581/0.3586&25.12/24.86/25.03\\
			
			Tree-Ring\cite{wen2023tree}&0.3543/0.3586/0.3602&24.83/24.87/25.13\\
			
			StableSignature$_{\textrm{[ICCV2023]}}$\cite{fernandez2023stable}&0.3541/0.3579/0.3599&24.59/24.72/25.07\\
			
			GaussianShading$_{\textrm{[CVPR2024]}}$\cite{yang2024gaussian}&0.3559/0.3594/0.3610&24.63/\textbf{24.36}/24.48\\
			
			PRCW$_{\textrm{[ICLR2025]}}$\cite{gunn2024undetectable}&0.3547/0.3581/0.3575&24.67/24.79/24.68\\
			
			
			\textbf{Ours}&0.3557/0.3588/0.3604&24.62/24.41/24.42\\
			
			\bottomrule
		\end{tabular}%
	}
	\caption{Comparisons of image quality on different methods.}
	\label{tab:tab2}
\end{table}

\begin{table}[h]

\centering
\setlength{\abovecaptionskip}{5pt}  
\setlength{\belowcaptionskip}{-10pt} 
\adjustbox{max width=1\linewidth}{%

\begin{tabular}{l|c|c|c}
\toprule

Samplers &TPR@F./Noise&BitAcc./Noise&TPR@B./Noise\\

\midrule

DDIM&1.000/1.000&1.0000/0.9925&1.000/0.965\\

DPMSolver&1.000/1.000&0.9936/0.9638&0.979/0.895\\

UniPC&1.000/1.000&0.9942/0.9685&0.982/0.899\\

PNDM&1.000/1.000&1.0000/0.9882&1.000/0.947\\

DEIS&1.000/1.000&1.0000/0.9956&1.000/0.988\\

Euler&1.000/1.000&1.0000/0.9671&0.983/0.893\\

\bottomrule
\end{tabular}%
}
\caption{Comparisons of the three metrics under different samplers}
\label{tab:tab3}
\end{table}

\noindent\textbf{Guidance Scales (CFG).} CFG is a technique used to enhance the correlation between the generated samples and the given conditions (such as text descriptions). It is also one of the sources of error in diffusion models. A higher CFG makes the model more faithful to the prompt, but increases the difficulty of reconstructing the initial noise during the inversion process. As shown in Fig.\ref{fig:fig7i}, while the voting rate increases with the CFG value, the TPR remains high with no degradation in performance.


\subsection{Baseline Comparison}

We compare the performance of the Gaussian-Shannon with other baseline methods. As shown in Tab.\ref{tab:tab1}, our method achieves results comparable to previous approaches under average noise levels. Furthermore, it demonstrates a significant advantage in scenarios requiring high precision.

\subsection{Ablation Study}

To systematically evaluate our method, we provide an empirical analysis of our design choices in this section.

\noindent\textbf{Code Rate \(R\).}The code rate refers to the proportion of useful information bits in the total transmitted bits. In theory, a low code rate provides stronger anti-interference capability. We selected five different code rates for ablation studies. To choose appropriate parameters, we did not fix the redundancy; instead, for each code rate, we selected the maximum redundancy while ensuring the latent-space capacity was not exceeded. As shown in Tab.\ref{tab:tab4}, a code rate of 1/4 exhibits the best error-correction capability. When the code rate is higher than 1/4, insufficient redundancy leads to degraded error correction; when the code rate is lower than 1/4, structural defects in the parity-check matrix of regular LDPC codes become pronounced, causing decoding failures. The response to this situation—carefully configuring an irregular LDPC code—is left for future work.

\begin{table}[h]
\centering
\setlength{\abovecaptionskip}{5pt}  
\setlength{\belowcaptionskip}{-15pt} 
\adjustbox{max width=1\linewidth}{
\begin{tabular}{lccccc}
\toprule

\multirow{2}{*}{Noise}

&\multicolumn{5}{c}{CodeRate}\\

\cmidrule(lr){2-6}

&1/6&1/5&1/4&1/3&1/2\\

\midrule

None&1.000/0.159&0.999/0.101&1.000/0.028&1.000/0.260&1.000/0.998\\

Noise&0.781/0.957&0.873/0.921&0.965/0.786&0.852/1.000&0.795/1.000\\

\bottomrule
\end{tabular}%
}
\caption{TPR@BitAcc.100\% and voting rate under different code rates}
\label{tab:tab4}
\end{table}

\noindent\textbf{Redundancy \(m\).} Redundancy refers to the number of repetitions of a codeword; the higher the count, the stronger the error-tolerance capability.  
As shown in Tab.\ref{tab:tab5}, under the highest redundancy ($m=16$) the number of voting rounds is minimal, implying that the codeword can correct errors autonomously.  
As redundancy gradually decreases, the required voting rounds increase.  
When there is no redundancy ($m=1$), voting becomes impossible and TPR drops below $100\%$, demonstrating that the error-correction method performs best when combined with redundancy.

\begin{table}[h]
\centering
\setlength{\abovecaptionskip}{5pt}  
\setlength{\belowcaptionskip}{-10pt} 
\adjustbox{max width=1\linewidth}{

\begin{tabular}{lccccc}
\toprule

\multirow{2}{*}{Noise}

&\multicolumn{5}{c}{Redundancy}\\

\cmidrule(lr){2-6}

&16&8&4&2&1\\

\midrule

None&1.000/0.028&1.000/0.021&1.000/0.028&1.000/0.040&0.929/0.071\\

Noise&0.965/0.786&0.739/1.000&0.592/1.000&0.314/1.000&0.187/0.813\\

\bottomrule
\end{tabular}%
}
\caption{TPR@BitAcc.100\% and voting rate under different redundancy levels}
\label{tab:tab5}
\end{table}

\noindent\textbf{SNR.} LDPC code decoding requires a prior estimate of the channel's signal-to-noise ratio (SNR). Over- or underestimating the environmental SNR will affect decoding accuracy. Generally, the lower the channel quality, the smaller the tolerable error. We set the initial SNR estimate of the channel to 0. As shown in the Tab.\ref{tab:tab6}, in a noise-free environment, performance is best and the voting rate is low when SNR is $0\sim5$, indicating that the true SNR lies within this interval. As the predicted environmental SNR deviates (SNR $<$ 0 or $>$ 5), performance gradually degrades. In average noise environments, the voting rate is high, indicating that the SNR exceeds the capability of the LDPC code. Majority voting is needed to improve codeword quality before error correction. After voting, TPR is better when SNR is $-2\sim2$. In summary, during decoding, we fix the environmental SNR of the entire channel at 0 dB, because in noise-free or post-voting conditions, an estimated SNR of 0 dB will not severely deviate from the true SNR.

\begin{table}[h]
\centering
\setlength{\abovecaptionskip}{5pt}  
\setlength{\belowcaptionskip}{-10pt} 
\adjustbox{max width=1\linewidth}{%
\begin{tabular}{lccccccc}
\toprule

\multirow{2}{*}{Noise}

&\multicolumn{7}{c}{SNR}\\

\cmidrule(lr){2-8}

&-10&-5&-2&0&2&5&10\\

\midrule

None&0.859/1.000&0.989/0.488&1.000/0.104&1.000/0.028&1.000/0.025&1.000/0.028&0.992/0.207\\

Noise&0.095/1.000&0.239/1.000&0.896/0.802&0.965/0.786&0.875/0.729&0.855/0.770&0.334/0.989\\
\bottomrule
\end{tabular}%
}
\caption{TPR@BitAcc.100\% and voting rate under different SNR}
\label{tab:tab6}
\end{table}

\subsection{Advanced Attacks}

This section evaluates the robustness of Gaussian Shannon under advanced attacks, compared with the methods StableSignature, Gaussian Shading, and PRCW. Following previous research, three types of advanced attacks are selected for experiments.

\noindent\textbf{Compression Attack.} We select two pre-trained VAE compressors as VAE1\cite{cheng2020learned} and VAE2\cite{begaint2020compressai} for image compression, and investigate the watermark quality after neural network compression.

\noindent\textbf{Diffusion Regeneration.} Regeneration attacks alter an image’s latent representation by first introducing noise and then applying a denoising process. Following a recent benchmark\cite{an2024waves}, we use a diffusion model pre-trained on ImageNet and perform 100 diffusion steps to attack the image. 

\noindent\textbf{Embedding Attack.}
Assuming the VAE model used by the original diffusion model is known, adversarial perturbations are applied to the image's embedding space. Specifically, we use the PGD algorithm\cite{madry2018towards} to generate adversarial images whose features differ significantly in latent space while having minimal impact in pixel space.

As shown in Tab.\ref{tab:tab7}, our method still shows strong robustness under the four attacks.  
Although TPR@BitAcc.100\% is only moderate under VAE2 and diffusion attacks, its overall performance outperforms other baselines.

\begin{table}[h]
\centering
\setlength{\abovecaptionskip}{5pt}  
\setlength{\belowcaptionskip}{-10pt} 
\adjustbox{max width=1\linewidth}{%
\begin{tabular}{l|c|ccccc}
\toprule
\textbf{Methods} & \textbf{DM} & \textbf{VAE1} & \textbf{VAE2} & \textbf{Diffusion} & \textbf{Embedding} \\
\midrule
Stable Signature & \multirow{4}{*}{\rotatebox[origin=c]{90}{SD V1.4}}  & 0.39/0.62/0.00&0.06/0.51/0.00 & 0.26/0.58/0.00 & 1.00/0.98/0.55\\
Gaussian Shading& & 1.00/0.99/0.26 & \textbf{0.98}/0.92/0.06&\textbf{0.98}/0.92/0.08&1.00/0.97/0.33\\
PRCW & &0.70/0.86/0.70 & 0.15/0.63/0.10 &  0.16/0.62/0.15 & 0.78/0.89/0.75\\
\textbf{Ours} & & \textbf{1.00/0.99/0.96} & 0.95/\textbf{0.92/0.68} & 0.96/\textbf{0.92/0.67}  & \textbf{1.00/0.98/0.92}\\
\midrule
Stable Signature & \multirow{4}{*}{\rotatebox[origin=c]{90}{SD V2.0}}  & 0.41/0.63/0.00 &0.06/0.51/0.00& 0.26/0.57/0.00 & 1.00/0.97/0.52  \\
Gaussian Shading& & 1.00/0.97/0.22 & \textbf{0.98}/0.91/0.08&\textbf{0.97}/0.91/0.09&1.00/0.97/0.34\\
PRCW & &0.67/0.85/0.67 & 0.16/0.63/0.08 &  0.14/0.59/0.15 & 0.78/0.88/0.75\\
\textbf{Ours} & & \textbf{1.00/0.99/0.95} & 0.95/\textbf{0.91/0.69} & 0.96/\textbf{0.92/0.67}  & \textbf{1.00/0.98/0.91}\\
\midrule
Stable Signature & \multirow{4}{*}{\rotatebox[origin=c]{90}{SD V2.1}} & 0.43/0.61/0.00 & 0.06/0.50/0.00  & 0.23/0.58/0.00 & 1.00/0.98/0.54 \\
Gaussian Shading& & 1.00/0.97/0.23 & \textbf{0.98}/0.91/0.07&\textbf{0.98}/0.91/0.08&1.00/0.97/0.35\\
PRCW & &0.70/0.85/0.70 & 0.12/0.56/0.12 & 0.14/0.57/0.14 & 0.76/0.88/0.76\\
\textbf{Ours} & & \textbf{1.00/0.99/0.95} & 0.94/\textbf{0.91/0.68} & 0.95/\textbf{0.91/0.65} & \textbf{1.00/0.98/0.92}\\
\bottomrule
\end{tabular}
}
\caption{Performance under four advanced attacks, we evaluate TPR@\(10^{-6}\)FPR, BitAcc. and TPR@BitAcc.100\%.}
\label{tab:tab7}
\end{table}

\section{Limitations}
\label{sec:formatting}


Our method has several limitations. First, under excessively high-intensity noise, the True Positive Rate drops significantly, which is primarily dictated by the thresholds in the majority voting and LDPC mechanisms. Second, as a semantic watermarking approach, the robustness of our method hinges on the preservation of meaningful image content. Consequently, the method may fail when severe geometric distortions substantially alter key visual features.
\section{Conclusion}
\label{sec:formatting}



We propose a communication-based watermarking framework for diffusion models. By modeling embedding and extraction as a reliable communication process, we unify robust watermark tracking and lossless recovery. Unlike previous methods, our approach guarantees bit-level accuracy under various noises while maintaining high image quality. Experiments show that our method achieves state-of-the-art performance across multiple diffusion variants, enabling practical offline verification and copyright provenance.

\section*{Acknowledgment}
\label{sec:formatting}

This work was supported by the Guangdong Provincial Key Fields Special Project for Ordinary Universities (2025ZDZX1027).

{
    \small
    \bibliographystyle{ieeenat_fullname}
    \bibliography{main}
}


\end{document}